\documentclass[10pt,twocolumn,letterpaper]{article}

\usepackage[pagenumbers]{cvpr} % To force page numbers, e.g. for an arXiv version

\usepackage{graphicx}
\usepackage{amsmath}
\usepackage{amssymb}
\usepackage{booktabs}

\usepackage[pagebackref,breaklinks,colorlinks]{hyperref}

\usepackage[capitalize]{cleveref}
\crefname{section}{Sec.}{Secs.}
\Crefname{section}{Section}{Sections}
\Crefname{table}{Table}{Tables}
\crefname{table}{Tab.}{Tabs.}

\def\cvprPaperID{*****} % *** Enter the CVPR Paper ID here
\def\confName{CVPR}
\def\confYear{2024}

\begin{document}

\title{Fashion Image Retrieval with Multi-Granular Alignment}

% \author{
% Jinkuan Zhu\textsuperscript{1}
% {\tt\small jinkuanzhu0@gmail.com}
% \and
% Hao Huang\\
% Kuaishou Inc.\\
% \and 
% Qiao Deng\\
% Kuaishou Inc.\\
% University of Electronic Science and Technology of China
% }
\author{
Jinkuan Zhu\textsuperscript{1}
\and
Hao Huang\textsuperscript{1}
\and
Qiao Deng\textsuperscript{1} 
\and 
Xiyao Li\textsuperscript{1} 
\and 
% \textsuperscript{1} Center for Future Media, University of Electronic Science and Technology of China 
\textsuperscript{1} Kuaishou Inc. \\
{\tt\small 
\{zhujinkuan, huanghao07, dengqiao, lixiyao\}@kuaishou.com
}
}
\maketitle

%%%%%%%%% ABSTRACT
\begin{abstract}
Fashion image retrieval task aims to search relevant clothing items of a query image from the gallery. The previous recipes focus on designing different distance-based loss functions, pulling relevant pairs to be close and pushing irrelevant images apart. However, these methods ignore fine-grained features (e.g. neckband, cuff) of clothing images. In this paper, we propose a novel fashion image retrieval method leveraging both global and fine-grained features, dubbed Multi-Granular Alignment (MGA). Specifically, we design a Fine-Granular Aggregator (FGA) to capture and aggregate detailed patterns. Then we propose Attention-based Token Alignment (ATA) to align image features at the multi-granular level in a coarse-to-fine manner. To prove the effectiveness of our proposed method, we conduct experiments on two sub-tasks (In-Shop \& Consumer2Shop) of the public fashion datasets DeepFashion. The experimental results show that our MGA outperforms the state-of-the-art methods by $1.8\%$ and $0.6\%$ in the two sub-tasks on the R@1 metric, respectively.
\end{abstract}

%%%%%%%%% BODY TEXT

\section{Introduction}

\begin{figure}[t]
  \centering
  \includegraphics[width=1.0\linewidth]{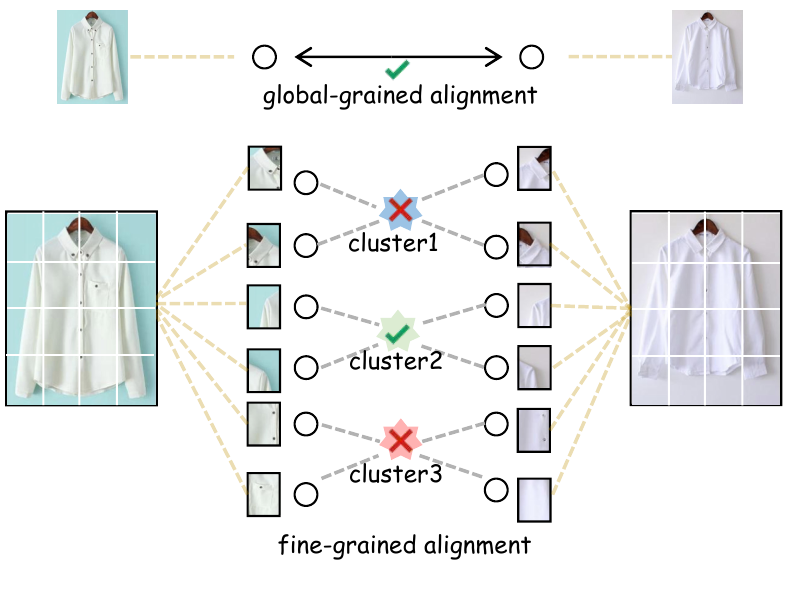}
  \caption{The motivation of our multi-granular alignment. Conventional methods measure the relevance between images by computing the similarity of the global features (the upper part). Our method further considers the fine-grained similarity (the lower part), which is achieved by an aggregating method FGA and an alignment method ATA.}
  \label{fig1}
\end{figure} 

\begin{figure*}
    \centering
    \includegraphics[scale=0.9]{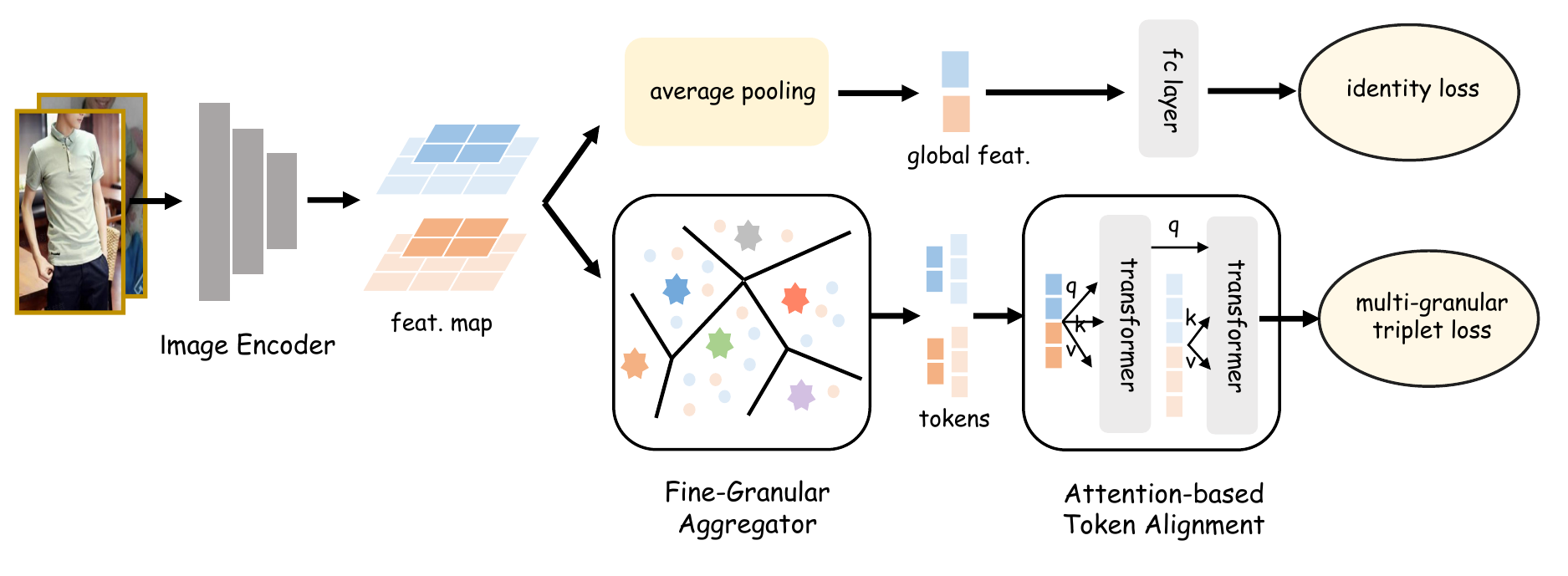}
    \caption{The overview of Multi-Granular Alignment framework. It mainly consists of three blocks: an image encoder to extract grid features from images. A fine-granular aggregator to cluster grid features into visual tokens. An attention-based token alignment block to compute the token-level similarity of images.}
    \label{fig2}
    % \vspace{-1.5em}
\end{figure*}

\label{sec:intro}
Image retrieval is an old but popular research topic that targets finding similar images from a database containing a big amount of images. With the rapid development of various Social Media and E-commerce platforms. Fashion image retrieval \cite{deepfashion} has been flavored by more and more researchers. Different from conventional image retrieval tasks (e.g. buildings), fashion images pay more attention to distinct information about clothes (e.g. texture, designs). Thus fashion image retrieval is challenging and worth exploring. 
\par Existing methods \cite{ctl, rst, grnet, hyp-vit, proxy-anchor, proxynca++, itir, nsoftmax, heejae, xbm} of fashion image retrieval follow the pipelines of image retrieval, which mainly have two components: feature extraction and distance-based loss functions. For feature extraction, some works modify CNN backbones for mining more visual information. HeeJae \etal \cite{heejae} combine three branches of feature extraction by several global descriptors, obtaining the effectiveness of an end-to-end ensemble manner. As for designing distance-based loss functions. Many pieces of research are proposed to improve the traditional triplet loss. Xun \etal \cite{xbm} propose cross-batch memory to enlarge the batch size during the training procedure, mining more hard negative samples for robust representation learning. Mikolaj \etal \cite{ctl} propose a centroid-based triplet loss to pull each image and a learned centroid vector which represents a center belonging to each class. Aleksandr \etal \cite{hyp-vit} propose a hyperbolic-based model to encode images and optimize the model in hyperbolic space. 
\par However, the aforementioned methods ignore the distinct character of fashion images. Different from the traditional image retrieval tasks (e.g. landmark, vehicle, product retrieval), fashion images are mostly clothing images collected from online E-commerce platforms. Thus these images usually have similar global characters (e.g. shape and style). As shown in \cref{fig1}, straightly representing a fashion image by a global feature may make it hard for the retrieval model to distinguish the clothing images. As human beings often distinguish fashion images by detailed information (e.g. designs of the neckband, patterns of logos) from clothes. How to mine fine-grained information from fashion images is worth exploring. 
\par Based on the above insights, in this paper, we try to leverage such fine-grained information and propose a novel fashion image retrieval model with Multi-Granular Alignment (MGA). Our model mainly has two novel designs: a Fine-Granular Aggregator (FGA) to extract fine-grained information and an Attention-based Token Alignment (ATA) to achieve multi-granular alignment. Specifically, for our FGA module, we first split the image feature maps at different layers into grid features for capturing detailed information for various scales. Then we aggregate the similar grid features and obtain a set of fine-granular tokens for each layer. As for the ATA module, we utilize tokens in the coarse feature map to attend tokens in the fined feature map by several transformer blocks. Depending on this progressive attention strategy, we obtain attended tokens containing detail information at different scales. Finally we compute the overall similarity by fusing global-granular and fine-granular similarity, where the fine-granular similarity is computed by the obtained attended tokens. Then we utilize a rank loss function to force relevant image pairs to be closed and irrelevant pairs to be apart. The core contributions of our paper are the following:
\begin{itemize}
    \item We propose a novel fashion image retrieval method that focus on capturing the multi-granular characters of clothing images and learning multi-granular representations by the global-to-fined alignment.
    \item We design a Fine-Granular Aggregator (FGA) to capture and aggregate fine-grained image features. An Attention-based Token Alignment (ATA) is introduced to align images with the detailed information. 
    \item We conduct experiments on two challenging fashion image retrieval benchmarks (In-Shop \& Consumer2Shop) using the public dataset DeepFashion \cite{deepfashion}. The experimental results demonstrate the superiority of our proposed method. 
\end{itemize}

%------------------------------------------------------------------------
\section{Method}
\label{sec:method}
The overview framework of our proposed MGA is depicted in Figure 1. The remainder of this section is organized as follows. In \cref{2.1}, we illustrate the feature extraction process and introduce our proposed Fine-Granular Aggregator module. In \cref{2.2}, we show our proposed multi-granular alignment. In \cref{2.3}, we demonstrate the overall training objective and loss functions.

\subsection{Feature Extraction}
\label{2.1}
\par \textbf{Image Encoder.} We utilize EfficientNetV2 \cite{effecientnetv2} pre-trained on ImageNet \cite{imagenet} to extract image features. Given an image $I$, we first split the feature maps $\Psi(I)$ at different backbone layers into the grid features $X^{grid} = \{x_{1}, x_{2}, ..., x_{n}\}$. Then we add an average pooling layer to the feature map at the final layer, obtaining the final global image features $X^{g} \in \mathbb{R}^{D}$, where $D$ is the dimension.

% before the final average pooling layer $\Psi(I) \in \mathbb{R}^{w \times h \times D}$ as the grid features $X^{grid} = \{x_{1}, x_{2}, ..., x_{n}\}$, $n=w \times\ h$. Then we add an average pooling layer to $\Psi(I)$, obtaining the final global image features $X^{g} \in \mathbb{R}^{D}$.   

\par \textbf{Fine-Granular Aggregator.}
The obtained grid features consist of various kinds of information about the image, including foreground and background, main body and details. To obtain informative fine-grained features, we propose our FGA module to aggregate grid features. Specifically, we design $M$ learnable vectors as $C=\{c_{1}, c_{2}, ..., c_{M}\}$ as cluster centers. Each center is capable to aggregate similar grid features and generate a representative feature (dubbed token). 

We first calculate the similarity between grid features $X^{grid} = \{x_{1}, x_{2}, ..., x_{n}\}$ and clusters $C$ by cosine similarity. Given each grid feature $x_{i}$, it can be mapped to the $j^{th}$ cluster in the following manner:
\begin{equation}
\label{eq1}
\gamma _{i,j} = \frac{exp(x_{i}c_{j}^{T} + b_{j})}{\sum_{k=1}^{M}{exp(x_{i}x_{k}^{T} + b_{k})}} 
\end{equation}
where $b_{k}$ is a trainable parameter. Then we obtain the token $T = \{t_{1}, t_{2}, ..., t_{M}\}$ of each center by a weighted summation of all grid features:  
\begin{equation}
\label{eq2}
t_{j} = Norm(\sum_{i=1}^{N}{\gamma_{i,j}(x_{i} - c_{j}^{*})})
\end{equation}
where "Norm" denotes $l_{2}$-normalization operation and $c_{j}^{*}$ is a learnable parameter which has the same size as $c_{j}$. 

\subsection{Multi-Granular Alignment}
\label{2.2}
We expect the extracted features of an image as a global-grained feature $X^{g}$ and a set of fine-grained tokens  $T = \{t_{1}, t_{2}, ..., t_{M}\}$. Multi-granular alignment aims to compute the similarity between images at both the global level and the token level. At the global level, we simply obtain the similarity by the cosine distance:
\begin{equation}
\label{eq3}
    S^{global} = \frac{X^{g}_{1}{X^{g}_{2}}^{T}}{|X^{g}_{1}||X^{g}_{2}|}
\end{equation}
As for the token level, we propose an attention-based method ATA to obtain the token-level similarity. 
\par \textbf{Attention-based Token Alignment.}
We define two stages to measure the similarity between images at the token level. Firstly, we send utilize several transformer blocks to obtain attended tokens, considering both collinear parts of pairs and coarse-to-fine details.  Secondly, we compute token-level similarity by the obtained attended tokens. 
\par In the first stage, we concatenate pair-wise tokens at a coarse layer and send them in to a transformer block. In this block, tokens of one image will attend tokens of the other to activate their similar parts. Then we send the output of this block as query into another transformer block, where the key and value are tokens at a finer layer. In this block, coarse tokens attend fined tokens for more detailed information. Depending on this progressive attention strategy, we obtain a set of attended tokens for each image. 
\par In the second stage, we use each attended token of one image to attend all tokens of the other image, obtaining a corresponding attended vector. Then we compare each token of one image with its corresponding vector, determining the relevance of this token with respect to the whole sequence of the other image. Then we average the relevance computed by all tokens to obtain the token-level similarity.
\par Specifically, given two token sequences $T^{v}=\{t^{v}_{1}, t^{v}_{2}, ..., t^{v}_{M}\}$ and $T^{e}=\{t^{e}_{1}, t^{e}_{2}, ..., t^{e}_{M}\}$, we first compute the cosine similarity matrix for all pairs:
\begin{equation}
\label{eq4}
    s_{i,j} = \frac{t^{v}_{i} {t^{e}_{j}}^{T}}{|t^{v}_{i}||t^{e}_{j}|}
\end{equation}
where $s_{i,j}$ represents the similarity between the $i^{th}$ token of $T^{v}$ and $j^{th}$ token of $T^{e}$. To obtain the corresponding attended vector with respect to each token $t^{v}_{i}$, we utilize a weighted combination of tokens from $T^{e}$:
\begin{equation}
\label{eq5}
\begin{aligned}
a^{e}_{i} &= \sum^{M}_{j=1}{w_{ij}t^{e}_{j}}\\
w_{ij} &= \frac{exp(\tau s_{ij})}{\sum_{j=1}^{M}exp(\tau s_{ij})}
\end{aligned}
\end{equation}
where $a^{e}_{i}$ is the attended vector with the respect to the $i^{th}$ token in $T^{v}$. And we can symmetrically obtain the attended vector $a^{v}_{i}$ as the corresponding attended vector of $t^{e}_{i}$ in the same way. $\tau$ is the temperature parameter of the softmax function. 
\par Then we determine the similarity at token-level by a maximizing operation of the relevance obtained by all tokens and their corresponding attended vectors:  
\begin{equation}
S^{token} = \mathop{\max}\limits_{i, i \in \{1,..,M\}} {\left [R(t^{v}_{i}, a^{e}_{i}) + R(t^{e}_{i}, a^{v}_{i}) \right ]} 
\end{equation}
where $R(t^{v}_{i}, a^{e}_{i})$ and $R(t^{e}_{i}, a^{v}_{i})$ represent the cosine similarity of pair $<t^{v}_{i}, a^{e}_{i}>$ and $<t^{e}_{i}, a^{v}_{i}>$, respectively. 

%------------------------------------------------------------------------
\subsection{Training}
\label{2.3}
In this section, we discuss the training objectives of our proposed MGA. We utilize two loss functions to optimize our model, which includes the common identity loss, center loss and our proposed multi-granular triplet loss. 
\par Following previous work, we regard the images belonging to the same clothing as one class and train a classification task through our model. Specifically, We add a $fc$ layer to the obtained global feature and the output is supervised by a common cross-entropy loss function, which we call identity loss. We also follow the previous work \cite{ctl} to add a center loss to pull together images of the same class.
As for the muti-granular triplet loss, we modified the conventional triplet loss $L = [\Delta - S(a, p) + S(a, n)]$, replacing the similarity between samples with a weighted combination of global-level similarity and token-level similarity: $S^{overall} = \alpha S^{global} + (1 - \alpha) S^{token}$, where $\alpha$ is a hyper-parameter. 

%------------------------------------------------------------------------
\section{Experiments}
\label{sec:experiments}
\subsection{Datasets}
\textbf{In-Shop clothes Retrieval} \cite{deepfashion} is a large subset of DeepFahsion, containing large pose and scale variations. It consists of $7,986$ categories of clothing items and $52,712$ number of In-shop clothes images in total. We utilize the first $3,997$ categories ($25,882$ images) for training. The remaining $3,997$ categories are partitioned into a query set ($14,218$ images) and a gallery set ($12,612$ images).  
\par \textbf{Consumer-to-Shop Retrieval} \cite{deepfashion} is also a large subset of Deepfahsion, which involves cross-domain correspondences and variations in the wild. It contains $3,3881$ categories of clothing items and $239,557$ number of consumer/shop clothes images. 
\subsection{Implementation Details}
We utilize EffecientNetV2 to extract image features in our method. We use Adam optimizer with a base learning rate of $1e^{-4}$ and multistep learning rate scheduler, decreasing the learning rate by a factor of 10 after $40^{th}$ and $70^{th}$ epochs. We train our model in a single V100 GPU. The batch size is set to 48 and the entire training procedure lasts 130 epochs. 
\par To evaluate the model performance, we report the R@K-acc and mAP metrics for comparison. These metrics reflect the quality of the results of a search engine as they would be visually inspected by a user. 

\subsection{Results}

To demonstrate the superiority of our MGA, we conduct experiments on two fashion image retrieval benchmarks and compare our model with the latest methods. The evaluation results are reported in \cref{tab1}. We can observe that our method outperforms the state-of-the-art methods on all the metrics. Especially noticeable is the increase of the R@1 metric, where our method outperforms the best competitor Hyp-ViT by $1.8\%$. It indicates our method improves the accuracy of the top-1 retrieving items. We believe it is benefited from our proposed multi-granular alignment, which leverages token-level information from images for both global-grained and fine-grained image comparison.

\begin{table}[]
\resizebox{\columnwidth}{!}{%
\begin{tabular}{lccccc}
\hline
\multicolumn{1}{c}{Model} & mAP           & R@1           & R@10          & R@20          & R@50          \\ \hline
\multicolumn{6}{c}{In-Shop}                                                                               \\ \hline
Proxy-Anchor \cite{proxy-anchor}              &               & 91.5          & 98.1          & 98.8          & 99.1          \\
NSoftmax \cite{nsoftmax}                  &               & 86.8          & 97.5          & 98.4          & 98.8          \\
ProxyNCA++ \cite{proxynca++}               &               & 90.4          & 98.1          & 98.8          & 99.0          \\
$IRT_{R}$ \cite{itir}                      &               & 91.9          & 98.1          & 98.7          & 98.9          \\
Hyp-ViT \cite{hyp-vit}                  &               & 92.5          & 98.3          & 98.8          & 99.1          \\ \hline
Ours                      & \textbf{81.1} & \textbf{94.3} & \textbf{98.8} & \textbf{99.1} & \textbf{99.5} \\ \hline
% Ours^{*}                      & \textbf{81.1} & \textbf{94.3} & \textbf{98.8} & \textbf{99.1} & \textbf{99.5} \\ \hline
\multicolumn{6}{c}{Consumer-to-Shop}                                                                      \\ \hline
GRNet \cite{grnet}                    &               & 25.7          &               & 64.4          & 75.0          \\
RST \cite{rst}                      & 43.0          & 37.8          & 71.1          & 77.2          & 84.1          \\
CTL \cite{ctl}                      & 43.1          & 37.6          & 71.1          & 77.6          & 84.7          \\ 
CTL+CE \cite{ctl}                      & 49.2          & 37.3          & 71.2          & 77.7          & 85.0          \\ \hline
Ours                      & 44.2 & \textbf{38.4} & 72.3 & 78.3 & 85.0 \\ 
Ours+CE                      & \textbf{50.3} & 38.1 & \textbf{72.7} & \textbf{78.8} & \textbf{85.4} \\ \hline
\end{tabular}%
}
\caption{Comparison with state-of-the-art methods. We compare our MGA with the latest fashion retrieval methods on In-shop and Consumer-to-Shop benchmarks. We evaluate the performance on 5 metrics: mAP, R@1, R@10, R@20 and R@50. 'CE' is a post-processing method proposed by \cite{ctl}. Higher is better for all the values in this table.}
\label{tab1}
\end{table}

\begin{table}[]
\resizebox{\columnwidth}{!}{%
\begin{tabular}{ccc|cccc}
\hline
baseline & \begin{tabular}[c]{@{}c@{}}normal\\ triplet loss\end{tabular} & \begin{tabular}[c]{@{}c@{}}multi-granular \\ triplet loss\end{tabular} & mAP  & R@1  & R@10 & R@20 \\ \hline
\checkmark       &                                                                         &                                                                        & 80.1  & 94.0  & 98.7  & 99.1  \\
\checkmark       & \checkmark                                                                      &                                                                        & 80.3 & 94.1 & 98.8 & 99.2 \\
\checkmark      &                                                                         & \checkmark                                                                    & 81.1 & 94.3 & 98.8 & 99.1 \\ \hline
\end{tabular}%
}
\caption{Ablation studies. We investigate the contribution of proposed modules and report the results on the In-Shop clothes Retrieval dataset.}
\label{tab2}
\end{table}

%------------------------------------------------------------------------
\subsection{Ablation Studies}
To investigate the contribution brought by our proposed components, we ablate our model into three configurations: 1) baseline model. we utilize the EffecientNetV2 to extract image features and add the identity loss and center loss to train the model; 2) normal triplet loss. We use features obtained by the baseline model to compute global-level similarity and feed it into a normal triplet loss to train the model; 3) multi-granular triplet loss. We utilize our proposed FGA module to extract token-level image features and the ATA module to compute the token-level similarity. Then we fuse global-level and token-level similarity to support the multi-granular triplet loss.

\par We conduct experiments on In-Shop clothes Retrieval dataset using the aforementioned three models. The results are depicted in \cref{tab2}. We have the following observations: 1) Adding a normal triplet loss only has a slight improvement  to the baseline model. It is because the normal triplet loss utilizes similarities computed by global image features, which is optimized by identity loss of the baseline model. Thus it is difficult to distinguish some hard-negative samples by detailed information. 2) Our proposed multi-granular triplet loss obviously boosts the performance of the baseline model. This benefit is owed to our proposed FGA and ATA modules. The FGA module captures the fine-grained image information and ATA module obtain token-level similarity considering the details of images which is hard to be distinguished.

%------------------------------------------------------------------------
\section{Conclusion}
In this paper, we introduce a novel fashion image retrieval method Multi-Granular Alignment (MGA), which considers similarity in a multi-grained way. We design a block FGA to aggregate fine-grained image features. We also propose a method ATA for progressively computing similarities at a fine-grained level. The results on two public benchmarks show the superiority of our model. And ablation studies prove the effectiveness of the proposed modules.

%%%%%%%%% REFERENCES
{\small
\bibliographystyle{ieee_fullname}
\bibliography{egbib}
}

\end{document}